%% file: main.tex
\documentclass[10pt,twocolumn,letterpaper]{article}
\usepackage[table,xcdraw,dvipsnames]{xcolor}
\definecolor{LightCyan}{rgb}{0.88,1,1}
\definecolor{LightYellow}{rgb}{1,1,0.7}
\definecolor{LightGreen}{rgb}{0.4, 1, 0.6}
\usepackage{cvpr}
\usepackage{times}
\usepackage{epsfig}
\usepackage{graphicx}
\usepackage{amsmath}
\usepackage{amssymb}
\usepackage{comment}
\usepackage{booktabs}
\usepackage{multirow}
\usepackage{multicol}
\usepackage{subfigure}
\usepackage{xr}
\usepackage{colortbl}

\usepackage[pagebackref=true,breaklinks=true,letterpaper=true,colorlinks,bookmarks=false]{hyperref}

\cvprfinalcopy 

\def\cvprPaperID{5256} 

\ifcvprfinal\fi
\begin{document}

\title{Learning monocular depth estimation infusing traditional stereo knowledge}

\author{Fabio Tosi, Filippo Aleotti, Matteo Poggi, Stefano Mattoccia\\
Department of Computer Science and Engineering (DISI)\\
University of Bologna, Italy\\
{\tt\small \{fabio.tosi5, filippo.aleotti2, m.poggi, stefano.mattoccia \}@unibo.it}
}

\maketitle

\begin{abstract}
Depth estimation from a single image represents a fascinating, yet challenging problem with countless applications. Recent works proved that this task could be learned without direct supervision from ground truth labels leveraging image synthesis on sequences or stereo pairs. Focusing on this second case, in this paper we leverage stereo matching in order to improve monocular depth estimation. To this aim we propose monoResMatch, a novel deep architecture designed to infer depth from a single input image by synthesizing features from a different point of view, horizontally aligned with the input image, performing stereo matching
between the two cues. In contrast to previous works sharing this rationale, our network is the first trained end-to-end from scratch.
Moreover, we show how obtaining proxy ground truth annotation through traditional stereo algorithms, such as Semi-Global Matching, enables more accurate monocular depth estimation still countering the need for expensive depth labels by keeping a self-supervised approach. Exhaustive experimental results prove how the synergy between i) the proposed monoResMatch architecture and ii) proxy-supervision attains state-of-the-art for self-supervised monocular depth estimation. The code is publicly available at \url{https://github.com/fabiotosi92/monoResMatch-Tensorflow}.

\end{abstract}

\input{figures/mrm_example.tex}

\section{Introduction}
Inferring accurate depth information of a sensed scene is paramount for several applications such as autonomous driving, augmented reality and robotics.  Although technologies such as LiDAR and time-of-flight are quite popular, obtaining depth from images is often the preferred choice. Compared to other sensors, those based on standard cameras potentially have several advantages: they are inexpensive, have a higher resolution and are suited for almost any environment. In this field, stereo is the preferred choice to infer \textit{disparity} (\ie, the inverse of depth) from two or more images sensing the same area from different points of view and \textit{Semi-Global Matching} (SGM) \cite{hirschmuller2005accurate} is a popular, yet effective algorithm to accomplish this task. However, inferring depth from a single image is particularly attractive because it does not require a stereo rig and overcomes some intrinsic limitations of a binocular setup (\eg, occlusions). On the other hand, it is an extremely challenging task due to the ill-posed nature of the problem. Nonetheless, deep learning enabled to achieve outstanding results for this task \cite{fu2018supervised}, although the gap with state-of-the-art stereo solutions is still huge \cite{Chang_2018_CVPR,liang2018learning}.
Self-supervised learning paradigms for monocular depth estimation \cite{monodepth17, zhou2017unsupervised, mahjourian2018unsupervised, 3net18, pydnet18, yang2018deep} became very popular to overcome the need for costly ground truth annotations, usually obtained employing expensive active sensors and human post-processing \cite{KITTI_2012,KITTI_2015,Uhrig2017THREEDV}. Following this strategy, Convolutional Neural Networks (CNNs) can be trained to tackle depth estimation as an image synthesis task from stereo pairs or monocular sequences \cite{monodepth17,zhou2017unsupervised}. For this purpose, using stereo pairs rather than monocular sequences as supervision turned out to be more effective according to the literature. Although the former strategy is more constrained since a stereo setup is necessary for training, it does neither require to infer relative pose between adjacent frames in a sequence nor to segment moving objects in the scene. Moreover, a stereo setup does not require camera motion, conversely to a monocular setup, to provide meaningful supervision. 
Other means for self-supervision consist into distilling \emph{proxy} labels in place of more expensive annotations for various tasks \cite{Tonioni_2017_ICCV,Tosi_2017_BMVC,DDFlow,makansi2018fusionnet,Klodt_2018_ECCV,guo2018learning}.

In this paper, we propose \underline{mono}cular \underline{Resi}dual \underline{Match}ing (shorten, monoResMatch), a novel end-to-end architecture trained to estimate depth from a monocular image leveraging a virtual stereo setup. 
In the first stage, we map input image into a features space, then we use such representation to estimate a first depth outcome and consequently synthesize features aligned with a \emph{virtual} right image. Finally, the last refinement module performs stereo matching between the real and synthesized representations.
Differently from other frameworks following a similar rationale \cite{luo2018single} that combines heterogeneous networks for synthesis \cite{xie2016deep3d} and stereo \cite{Mayer_2016_CVPR}, we use a single architecture trained in end-to-end fashion yielding a notable accuracy improvement compared to the existing solutions.
Moreover, we leverage traditional knowledge from stereo to obtain accurate proxy labels in order to improve monocular depth estimation supervised by stereo pairs. We will show that, despite the presence of outliers in the produced labels, training according to this paradigm results in superior accuracy compared to image warping approaches for self-supervision. 
Experimental results on the KITTI raw dataset \cite{KITTI_RAW} will show that the synergy between the two aforementioned key components of our pipeline enables to achieve state-of-the-art results compared to other self-supervised frameworks for monocular depth estimation not requiring any ground truth annotation.
Figure \ref{fig:abstract} shows an overview of our framework, depicting an input frame and the outcome of monoResMatch.

\section{Related Work}

In this section, we review the literature relevant to our work concerned with stereo/monocular depth estimation and proxy label distillation.

\textbf{Stereo depth estimation.}
Most conventional dense stereo algorithms rely on some or all the well-known four steps thoroughly described in \cite{scharstein2002taxonomy}. In this field, SGM \cite{hirschmuller2005accurate} stood out for the excellent trade-off between accuracy and efficiency thus becoming very popular.   
$\breve{Z}$bontar and LeCun \cite{zbontar2016stereo} were the first to apply deep learning to stereo vision replacing the conventional matching costs calculation with a siamese CNN network trained to predict the similarity between patches. Luo \etal \cite{luo2016efficient} cast the correspondence problem as a multi-class classification task, obtaining better results. Mayer \etal \cite{Mayer_2016_CVPR} backed away from the previous approaches and proposed an end-to-end trainable network called \textit{DispNetC} able to infer disparity directly from images. While DispNetC applies a 1-D correlation to mimic the cost volume, GCNet by Kendall \etal \cite{Kendall_2017_ICCV} exploited 3-D convolutions over a 4-D volume to obtain matching costs and finally applied a differentiable version of \textit{argmin} to select the best disparity along this volume. Other works followed these two main strategies, building more complex architectures starting from DispNetC \cite{Pang_2017_ICCV_Workshops,Liang_2018_CVPR,yang2018segstereo,song2018stereo} or GCNet \cite{Chang_2018_CVPR,yu2018kandao,khamis2018stereonet} respectively. The domain shift issue affecting these architectures (\eg synthetic to real) has been addressed in either offline \cite{Tonioni_2017_ICCV} or online \cite{Tonioni_2019_CVPR} fashion, or greatly reduced by guiding them with external depth measurements (\eg Lidar) \cite{POGGI_2019_CVPR}.

\textbf{Monocular depth estimation.}
Before the deep learning era, some works tackled depth-from-mono with MRF \cite{saxena2009make3d} or boosted classifiers \cite{ladicky2014pulling}. However, with the increasing availability of ground truth depth data, supervised approaches based on CNNs \cite{laina2016deeper,liu2016learning, xu2018supervised,fu2018supervised} rapidly outperformed previous techniques.
An attractive trend concerns the possibility of learning depth-from-mono in a self-supervised manner, avoiding the need for expensive ground truth depth labels that are replaced by multiple views of the sensed scene. Then, supervision signals can be obtained by image synthesis according to the estimated depth, camera pose or both.
In general, acquiring images from a stereo camera enables a more effective training than using a single, moving camera, since the pose between frames known.
Concerning stereo supervision, Garg \etal \cite{garg2016unsupervised} first followed this approach, while Godard \etal \cite{monodepth17} introduced spatial transform network \cite{jaderberg2015spatial} and a left-right consistency loss. Other methods improved efficiency \cite{pydnet18}, deploying a pyramidal architecture, and accuracy by simulating a trinocular setup \cite{3net18} or including joint semantic segmentation \cite{ramirez2018}. In \cite{DATE_2019}, a strategy was proposed to reduce further the energy efficiency of \cite{pydnet18} leveraging fixed-point quantization. The semi-supervised framework by Kuznietsov \etal \cite{Kuznietsov_2017_CVPR} combined stereo supervision with sparse LiDAR measurements.
The work by Zhou \etal \cite{zhou2017unsupervised} represents the first attempt to supervise a depth-from-mono framework with single camera sequences. This approach was improved including additional cues such as point-cloud alignment \cite{mahjourian2018unsupervised}, differentiable DVO \cite{wang2018unsupervised} and multi-task learning \cite{zou2018df}. Zhan \etal \cite{zhan2018unsupervised} combined the two supervision approaches outlined so far deploying stereo sequences.
Another class of methods \cite{atapour2018real, Aleotti_monogan_2018, kumar2018gan} applied a generative adversarial paradigm to the monocular scenario.

Finally, relevant to our work is Single View Stereo matching (SVS) \cite{luo2018single}, processing a single image to obtain a second synthetic view using Deep3D \cite{xie2016deep3d} and then computing a disparity map between the two using DispNetC \cite{Mayer_2016_CVPR}. However, these two architectures are trained independently. Moreover, DispNetC is supervised with ground truth labels from synthetic \cite{Mayer_2016_CVPR} and real domains \cite{KITTI_2015}. Differently, the framework we are going to introduce requires no ground truth at all and is elegantly trained in an end-to-end manner, outperforming SVS by a notable margin.

\input{figures/network.tex}
    
\textbf{Proxy labels distillation.} Since for most tasks ground truth labels are difficult and expensive to source, some works recently enquired about the possibility to replace them with easier to obtain proxy labels. Tonioni \etal \cite{Tonioni_2017_ICCV} proposed to adapt deep stereo networks to unseen environments leveraging traditional stereo algorithms and confidence measures \cite{Poggi_2017_ICCV}, Tosi \etal \cite{Tosi_2017_BMVC} learned confidence estimation selecting positive and negative matches by means of traditional confidence measures, Makansi \etal \cite{makansi2018fusionnet} and Liu \etal \cite{DDFlow} generated proxy labels for training optical flow networks using conventional methods.
Specifically relevant to monocular depth estimation are the works proposed by Yang \etal \cite{yang2018deep}, using stereo visual odometry to train monocular depth estimation, by Klodt and Vedaldi \cite{Klodt_2018_ECCV}, leveraging structure from motion algorithms and by Guo \etal \cite{guo2018learning}, obtaining labels from a deep network trained with supervision to infer disparity maps from stereo pairs.

\section{Monocular Residual Matching}

In this section, we describe in detail the proposed \emph{monocular Residual Matching} (monoResMatch) architecture designed to infer accurate and dense depth estimation in a self-supervised manner from a single image. Figure \ref{fig:monoResMatch} recaps the three key components of our network. First, a multi-scale feature extractor takes as input a single raw image and computes deep learnable representations at different scales from quarter resolution $F_L^2$ to full-resolution $F_L^0$ in order to toughen the network to ambiguities in photometric appearance. Second, deep high-dimensional features at input image resolution are processed to estimate, through an hourglass structure with skip-connections, multi-scale inverse depth (\ie, disparity) maps aligned with the input and a \emph{virtual} right view learned during training. By doing so, our network learns to emulate a binocular setup, thus allowing further processing in the stereo domain \cite{luo2018single}.  
Third, a disparity refinement stage estimates residual corrections to the initial disparity. In particular, we use deep features from the first stage and back-warped features of the \emph{virtual} right image to construct a cost volume that stores the stereo matching costs using a correlation layer \cite{Mayer_2016_CVPR}.

Our entire architecture is trained from scratch in an end-to-end manner, while SVS \cite{luo2018single} by training its two main components, Deep3D \cite{xie2016deep3d} and DispNetC \cite{Mayer_2016_CVPR}, on image synthesis and disparity estimation tasks separately (with the latter requiring additional, supervised depth labels from synthetic imagery \cite{Mayer_2016_CVPR}).

Extensive experimental results will prove that monoResMatch enables much more accurate estimations compared to SVS and other state-of-the-art approaches.

\subsection{Multi-scale feature extractor}

Inspired by \cite{Liang_2018_CVPR}, given one input image $I$ we generate deep representations using layers of convolutional filters. In particular, the first 2-stride layer convolves $I$ with 64 learnable filters of size $7\times7$ followed by a second 2-stride convolutional layer composed of 128 filters with kernel size $4\times4$. Two deconvolutional blocks, with stride 2 and 4, are deployed to upsample features from lower-spatial resolution to full input resolution producing 32 features maps each. A $1\times1$ convolutional layer with stride 1 further processes upsampled representations.

\subsection{Initial Disparity Estimation}

Given the features extracted by the first module, this component is in charge of estimating an initial disparity map. In particular, an encoder-decoder architecture inspired by DispNet processes deep features at quarter resolution from the multi-scale feature extractor (\ie, \textit{conv2}) and outputs disparity maps at different scales, specifically from $\frac{1}{128}$ to full-resolution. Each down-sampling module, composed of two convolutional blocks with stride 2 and 1 each, produces a growing number of extracted features, respectively 64, 128, 256, 512, 1024, and each convolutional layer uses $3\times3$ kernels followed by ReLU non-linearities. Differently from DispNet, which computes matching costs in the early part of this stage using features from the left and right images of a stereo pair, our architecture lacks such necessary information required to compute a cost volume since it processes a single input image.  Thus, no 1-D correlation layer can be imposed to encode geometrical constraints in this stage of our network. Then, upsampling modules are deployed to enrich feature representations through skip-connections and to extract two disparity maps, aligned respectively with the input frame and a \emph{virtual} viewpoint on its right as in \cite{monodepth17}. This process is carried out at each scale using 1-stride convolutional layers with kernel size $3\times3$.

\subsection{Disparity Refinement}
\label{sec:refinement}

Given an initial estimate of the disparity at each scale obtained in the second part of the network, often characterized by errors at depth discontinuities and occluded regions, this stage predicts corresponding multi-scale residual signals \cite{he2016deep} by a few stacked nonlinear layers that are then used to compute the final left-view aligned disparity map. This strategy allows us to simplify the end-to-end learning process of the entire network. Moreover, motivated by \cite{luo2018single}, we believe that geometrical constraints can play a central role in boosting the final depth accuracy. For this reason, we embed matching costs in feature space computed employing a horizontal correlation layer, typically deployed in deep stereo algorithms. To this end, we rely on the right-view disparity map computed previously to generate right-view features $\tilde{F}_R^0$ from the left ones $F_L^0$ using a differentiable bilinear sampler \cite{jaderberg2015spatial}. 
The network is also fed with error $e_L$, \ie the absolute difference between left and \textit{virtual} right features at input resolution, with the latter back-warped at the same coordinates of the former, as in \cite{liang2018learning}.

We point out once more that, differently from \cite{luo2018single}, our architecture produces both a synthetic right view, \ie its features representation, and computes the final disparity map following stereo rationale. This makes monoResMatch a single end-to-end architecture, effectively performing stereo out of a single input view rather than the combination of two models (\ie, Deep3D \cite{xie2016deep3d} and DispNetC \cite{Mayer_2016_CVPR} for the two tasks outlined) trained independently as in \cite{luo2018supervised}. Moreover, exhaustive experiments will highlight the superior accuracy achieved by our fully self-supervised, end-to-end approach.

\input{figures/proxy_example.tex}

\subsection{Training Loss}

In order to train our multi-stage architecture, we define the total loss as a sum of two main contributions, a $\mathcal{L}_{init}$ term from the initial disparity estimation module and a $\mathcal{L}_{ref}$ term from the disparity refinement stage. Following \cite{godard2018digging}, we embrace the idea to up-sample the predicted low-resolution disparity maps to the full input resolution and then compute the corresponding signals. This simple strategy is designed to force the inverse depth estimation to reproduce the same objective at each scale, thus leading to much better outcomes. In particular, we obtain the final training loss as:

\begin{equation} \label{eq:total_loss}
  \mathcal{L}_{total} = \sum_{s=1}^{n_i} \mathcal{L}_{init} + \sum_{s=1}^{n_r} \mathcal{L}_{ref}
\end{equation}

where $s$ indicates the output resolution, $n_i$ and $n_r$ the numbers of considered scales during loss computation, while $\mathcal{L}_{init}$ and $\mathcal{L}_{ref}$ are formalised as:

\input{equations/main_loss.tex} 

where $\mathcal{L}_{ap}$ is an image reconstruction loss, $\mathcal{L}_{ds}$ is a smoothness term and $\mathcal{L}_{ps}$ is a proxy-supervised loss. Each term contains both the left and right components for the initial disparity estimator, and the left components only for the refinement stage.

\smallskip
\textbf{Image reconstruction loss.}
A linear combination of $l_{1}$ loss and \textit{structural similarity measure} (SSIM) \cite{SSIM} encodes the quality of the reconstructed image $\tilde{I}$ with respect to the original image $I$:

\input{equations/reconstruction_loss.tex}

Following \cite{monodepth17}, we set $\alpha = 0.85$  and use a SSIM with $3 \times 3$ block filter.

\smallskip
\textbf{Disparity smoothness loss.}
This cost encourages the predicted disparity to be locally smooth. Disparity gradients are weighted by an edge-aware term from image domain:

\input{equations/smoothness_loss.tex}

\smallskip

\textbf{Proxy-supervised loss.}
Given the proxy disparity maps obtained by a conventional stereo algorithm, detailed in Section \ref{sec:distillation}, we coach the network using reverse Huber (berHu) loss \cite{owen2007robust}: 

\input{equations/berhu_loss.tex}

where $d_{ij}$ and $d_{ij}^{st}$ are, respectively, the predicted disparity and the proxy annotation for pixel at the coordinates $i,j$ of the image, while $c$ is adaptively set as $\alpha \max_{i,j}|d_{ij}-d_{ij}^{st}|$, with $\alpha = 0.2$.

\smallskip
\section{Proxy labels distillation}\label{sec:distillation}

To generate accurate proxy labels, we use the popular SGM algorithm \cite{hirschmuller2005accurate}, a fast yet effective solution to infer depth from a rectified stereo pair without training. 
In our implementation, initial matching costs are computed for each pixel $p$ and disparity hypothesis $d$ applying a $9\times7$ census transform and computing Hamming distance on pixel strings. Then, scanline optimization along eight different paths refines the initial cost volume as follows:

\begin{equation}\label{eq:sgm}
\begin{split}
E(p,d) = &C(p,d) + \min_{j>1}[C(p',d),C(p',d\pm1)+P1, \\ &C(p',d\pm q)+P2] -\min_{k<D_{max}}(C(p',k)) \\
\end{split}
\end{equation}
being $C(p,d)$ the matching cost for pixel $p$ and hypothesis $d$, $P_1$ and $P_2$ two smoothness penalties, discouraging disparity gaps between $p$ and previous pixel $p'$ along the scanline path. The final disparity map $D$ is obtained applying a winner-takes-all strategy to each pixel of the reference image. Although SGM generates quite accurate disparity labels, outliers may affect the training of a depth model negatively, as noticed by Tonioni \etal \cite{Tonioni_2017_ICCV}. They applied a learned confidence measure \cite{Poggi_2016_BMVC} to filter out erroneous labels when computing the loss. Differently, we run a non-learning based left-right consistency check to detect outliers. Purposely, by extracting both disparity maps $D^L$ and $D^R$ with SGM, respectively for the left and right images, we apply the following criteria to invalidate (\ie, set to -1) pixels having different disparities across the two maps:

\begin{equation}\label{eq:lrc}
    D(p)=
\begin{cases}
D(p) & \text{if } |D^L(p)-D^R(p-D^L(p))|\leq \varepsilon \\
-1 & \text{otherwise } \\
\end{cases}
\end{equation}

The left-right consistency check is a simple strategy that removes many wrong disparity assignments, mostly near depth discontinuities, without needing any training that would be required by \cite{Tonioni_2017_ICCV}. Therefore, our proxy labels generation process does not rely at all on ground truth depth labels. 
Figure \ref{fig:distillation} shows an example of distilled labels (b), where black pixels correspond to outliers filtered out by left-right consistency. Although some of them persist, we can notice how they do not affect the final prediction by the trained network and how our proposal can recover accurate disparity values in occluded regions on the left side of the image (c).

\section{Experimental results}\label{sec:experiments}

In this section, we describe the datasets, implementation details and then present exhaustive evaluations of monoResMatch on various training/testing configurations, showing that our proposal consistently outperforms self-supervised state-of-the-art approaches. As standard in this field, we assess the performance of monocular depth estimation techniques following the protocol by Eigen \etal \cite{eigen2014depth}, extracting data from the KITTI \cite{KITTI_RAW} dataset, using sparse LiDAR measurements as ground truth for evaluation. Additionally, we also perform an exhaustive ablation study proving that proxy supervision from SGM algorithm and effective architectural choices enable our strategy to improve predicted depth map accuracy by a large margin.

\input{tables/godard_comparison.tex}

\subsection{Datasets}

For all our experiments we compute standard monocular metrics \cite{eigen2014depth,monodepth17}: \textit{Abs rel}, \textit{Sq rel}, \textit{RMSE} and \textit{RMSE log} represent error measures while $\delta < \zeta$ the percentage of predictions whose maximum between ratio and inverse ratio with respect to the ground truth is lower than a threshold $\zeta$. Two main datasets are involved in our evaluation, that are KITTI \cite{KITTI_RAW} and CityScapes \cite{cordts2016cityscapes}.

\textbf{KITTI.}
The KITTI stereo dataset \cite{KITTI_RAW} is a collection of rectified stereo pairs made up of 61 scenes (containing about 42,382 stereo frames) mainly concerned with driving scenarios. Predominant image size is $1242 \times 375$ pixels. A LiDAR device, mounted and calibrated in proximity to the left camera, was deployed to measure depth information.

Following other works \cite{eigen2014depth, monodepth17}, we divided the overall dataset into two subsets, composed respectively of 29 and 32 scenes. We used 697 frames belonging to the first group for testing purposes and 22600 more taken from the second for training. We refer to these subsets as \textit{Eigen split}.

\textbf{CityScapes.}
The CityScapes dataset \cite{cordts2016cityscapes} contains stereo pairs concerning about 50 cities in Germany taken from a moving vehicle in various weather conditions. It consists of 22,973 stereo pairs with a shape of $2048 \times 1024$ pixels. Since most of the images include the hood of the car, mostly reflective and thus leading to wrong estimates, we discarded the lower $20\%$ of the frame before applying the random crop during training \cite{monodepth17}.

\subsection{Implementation details}

Following the standard protocol in this field, we used CityScapes followed by KITTI for training. We refer to these two training sets as Cityscapes (CS) and Eigen KITTI split (K) from now on. We implemented our architecture using the TensorFlow framework, counting approximately $42.5$ millions of parameters, summing variables from the multi-scale feature extractor (0.51 M), the initial disparity stage (41.4 M) and the refinement module (0.6 M). 
In the experiments, we pre-trained monoResMatch on CS running about 150k iteration using a batch size of 6 and random crops of size $512 \times 256$ on $1024 \times 512 $ resized images from the original resolution. We used Adam optimizer \cite{kingma2014adam} with $\beta_{1} = 0.9$, $\beta_{2} = 0.999$ and $\epsilon=10^{-8}$. We set the initial learning rate to $10^{-4}$, manually halved after $100$k and $120$k steps, then continuing until convergence. After the first pre-initialisation procedure, we perform fine-tuning of the overall architecture on 22600 KITTI raw images from K. Specifically, we run $300$k steps using a batch size of 6 and extracting random crops of size $640\times192$ from resized images at $1280\times384$ resolution. At this stage, we employed a learning rate of $10^{-4}$, halved after $180$k and $240$k iterations. We fixed the hyper-parameters of the different loss components to $\alpha_{ap}=1, \alpha_{ds}=0.1$ and $\alpha_{ps}=1$, while $n_i=4$ and $n_r=3$.
As in \cite{monodepth17}, data augmentation procedure has been applied to both images from CS and K at training, in order to increase the robustness of the network. At test time, we post-process disparity as in \cite{monodepth17,3net18,yang2018deep}. Nevertheless, we preliminary highlight that, differently from the strategies mentioned above, effects such as disparity ramps on the left border are effectively solved by simply picking random crops on proxy disparity maps generated by SGM, as clearly visible in Figure \ref{fig:distillation} (c).

\input{tables/overall_comparison.tex}
Proxy supervision is obtained through SGM implementation from \cite{spangenberg2014large}, which allows us to quickly generate disparity maps aligned with the left and right images for both CS and K. We process such outputs using left-right consistency check in order to reduce the numbers of outliers, as discussed in Section \ref{sec:distillation} using an $\epsilon$ of 1. We assess the accuracy of our proxy generator on 200 high-quality disparity maps from KITTI 2015 training dataset \cite{KITTI_2015}, measuring $96.1\%$ of pixels having disparity error smaller than 3. Compared to Tonioni \etal \cite{Tonioni_2017_ICCV}, we register a negligible drop in accuracy from 99.6\% reported in their paper. However, we do not rely on any learning-based confidence estimator as they do \cite{Poggi_2016_BMVC}, so we maintain label distillation detached from the need for ground truth as well.
Since SGM runs over images at full resolution while monoResMatch inputs are resized to $1280 \times 384$ before extracting crops, we enforce a scaling factor to SGM disparities given by $\frac{1280}{W}$, where $W$ is the original image \textit{width}. Consequently, the depth map estimated by monoResMatch must be properly multiplied by $\frac{W}{1280}$ at test time.  
The architecture is trained end-to-end on a single Titan XP GPU without any stage-wise procedure and infers depth maps in 0.16s per frame at test time, processing images at KITTI resolution (\ie, about $1280\times384$ to be compatible with monoResMatch downsampling factors).

\subsection{Ablation study}
In this section we examine the impact of i) proxy-supervision from SGM and ii) the different components of monoResMatch. The outcomes of these experiments, conducted on the Eigen split, are collected in Table \ref{tab:comparison_monodepth}. 

\textbf{Proxy-supervised loss analysis.} We train \emph{monodepth} framework by Godard \etal \cite{monodepth17} from scratch adding our proxy-loss, then we compare the obtained model with the original one, as well as with the more effective strategy used by 3Net \cite{3net18}. We can observe that proxy-loss enables a more accurate \emph{monodepth} model (row 3) compared to \cite{monodepth17}, moreover it also outperforms virtual trinocular supervision proposed in \cite{3net18}, attaining better metrics with respect of both, but $\delta<1.25$ for 3Net. 
Specifically, by recalling Figure \ref{fig:distillation}, the proxy distillation couples well with a cropping strategy, solving well-known issues for stereo supervision such as disparity ramps on the left border. 
We refer to supplementary material for additional qualitative examples.

\textbf{Component analysis.} Still referring to Table \ref{tab:comparison_monodepth}, we evaluate different configurations of our framework by ablating the key modules peculiar to our architecture. 
First, we train monoResMatch on K without proxy supervision (row 3) to highlight that our architecture already outperforms \cite{monodepth17} (row 1). 
Training on CS+K with proxy labels, we can notice how without any refinement module (\textit{no-refinement}), our framework already outperforms the proxy-supervised ResNet50 model of Godard \etal \cite{monodepth17}. Adding the disparity refinement component without encoding any matching relationship (\textit{no-corr}) enables small improvements, becoming much larger on most metrics when a correlation layer is introduced (\textit{no-pp}) to process real and synthesized features as to resemble stereo matching. Finally, post-processing as in \cite{monodepth17} (row 11) still ameliorates all scores, although the larger contribution is given by the correlation-based refinement module, as perceived by comparing \textit{no-refinement} and \textit{no-pp} entries. Finally, by comparing rows 4 and 11 we can also perceive the impact given by CS pretraining on our full model.

\input{tables/luo_comparison.tex}
\input{figures/kitti_benchmark.tex}
\input{tables/kitti_benchmark.tex}

\subsection{Comparison with self-supervised frameworks}

Having studied in detail the contribution of both monoResMatch architecture and proxy supervision, we compare our framework with state-of-the-art self-supervised approaches for monocular depth estimation.
Table \ref{table:eigen} collects results obtained evaluating different models on the aforementioned Eigen split \cite{eigen2014depth}. In this evaluation, we consider only competitors trained without \emph{any} supervision from ground truth labels (\eg, synthetic datasets \cite{Mayer_2016_CVPR}) involved in \emph{any} phase of the training process \cite{luo2018single,guo2018learning}. We refer to methods using monocular supervision (\textit{Seq}), binocular (\textit{Stereo}) or both (\textit{Seq+Stereo}). Most methods are trained on CS and K, except Yang \etal \cite{yang2018deep} that leverages on different sub-splits of K. From the table, we can notice that monoResMatch outperforms all of them significantly.

To compete with methods exploiting supervision from dense synthetic ground truth \cite{Mayer_2016_CVPR}, we run additional experiments using very few annotated samples from KITTI as in \cite{luo2018supervised,guo2018learning}, for a more fair comparison. Table \ref{table:eigen_200_500_700} collects the outcome of these experiments according to different degrees of supervision, in particular using accurate ground truth labels from the KITTI 2015 training split (200-acrt) or different amounts of samples from K with LiDAR measurements, respectively 100, 200, 500 and 700 as proposed in \cite{luo2018supervised,guo2018learning}, running only 5k iterations for each configuration. We point out that monoResMatch, on direct comparisons to methods trained with the same amount of labeled images, consistently achieves better scores, with rare exceptions. Moreover, we highlight in red for each metric the best score among all the considered configurations, figuring out that monoResMatch trained with 200-acrt plus 500 samples from K attains the best accuracy on all metrics. This fact points out the high effectiveness of the proposed architecture, able to outperform state-of-the-art techniques \cite{luo2018single,guo2018learning} trained with much more supervised data (\ie, more than 30k stereo pairs from \cite{Mayer_2016_CVPR} and pre-trained weights from ImageNet). Leveraging on the traditional SGM algorithm instead of a deep stereo network as in \cite{guo2018learning} for proxy-supervision ensures a faster and easier to handle training procedure.

\subsection{Performance on single view stereo estimation}

Finally, we further compare monoResMatch directly with Single View Stereo (SVS) by Luo \etal \cite{luo2018single}, being both driven by the same rationale. We fine-tuned monoResMatch on the KITTI 2015 training set as in Table \ref{table:eigen_200_500_700} and submitted to the online stereo benchmark \cite{KITTI_2015} as performed in \cite{luo2018supervised}. Table \ref{table:kitti-stereo} compares monoResMatch with SVS and other techniques evaluated in \cite{luo2018supervised}, respectively monodepth \cite{monodepth17} and OpenCV Block-Matching (OCV-BM). D1 scores represent the percentages of pixels having a disparity error larger than 3 or 5\% of the ground truth value on different portions of the image, respectively background (bg), foreground (fg) or its entirety (all). We can observe from the table a margin larger than 3\% on D1-bg and near to 1\% for D1-fg, resumed in a total reduction of 2.72\%.
This outcome supports once more the superiority of monoResMatch, although SVS is trained on many, synthetic images with ground truth \cite{Mayer_2016_CVPR}. 
Finally, Figure \ref{fig:kitti_test} depicts qualitative examples retrieved from the KITTI online benchmark.

\section{Conclusions}

In this paper, we proposed monoResMatch, a novel framework for monocular depth estimation. It combines i) pondered design choices to tackle depth-from-mono in analogy to stereo matching, thanks to a correlation-based refinement module and ii) a more robust self-supervised training leveraging on proxy ground truth labels generated through a traditional  (\ie non-learning based) algorithm such as SGM. 
In contrast to state-of-the-art models \cite{luo2018single,guo2018learning,yang2018deep}, our architecture is elegantly trained in an end-to-end manner.
Through exhaustive experiments, we prove that plugging proxy-supervision at training time leads to more accurate networks and, coupling this strategy with monoResMatch architecture, is state-of-the-art for self-supervised monocular depth estimation.

\textbf{Acknowledgement.} We gratefully acknowledge the support of NVIDIA Corporation with the donation of the Titan Xp GPU used for this research.

{
\small
\bibliographystyle{ieee}
\bibliography{bibliography}
}

\end{document}

%% file: figures/mrm_example.tex
\begin{figure}
\centering
\includegraphics[width=0.45\textwidth]{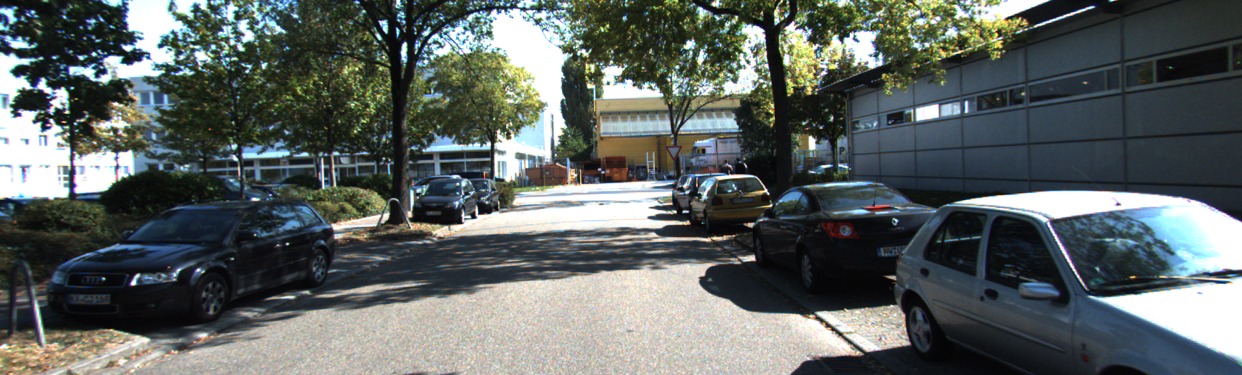} 
\includegraphics[width=0.45\textwidth]{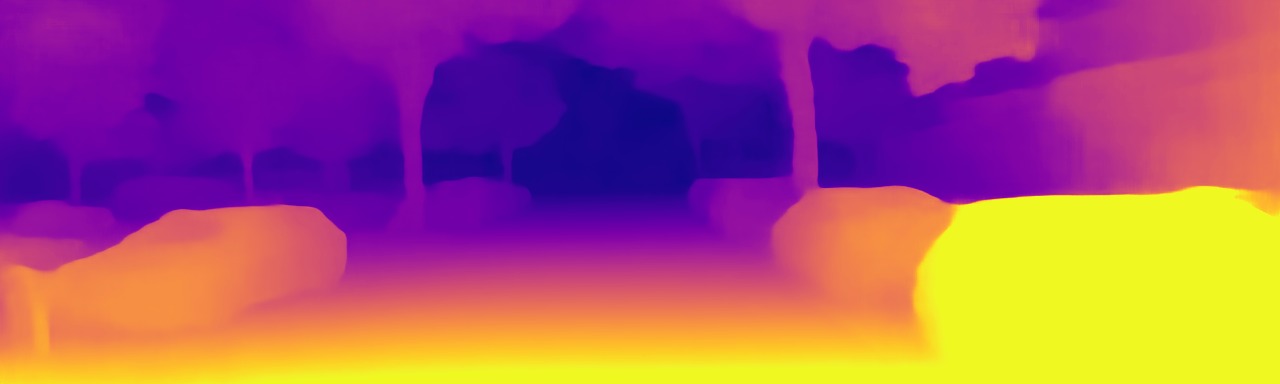} 
\caption{Overview of the proposed depth-from-mono solution. Input image from KITTI dataset (top). Estimated depth map by our monoResMatch (bottom).}
\label{fig:abstract}
\end{figure}

%% file: figures/network.tex
\begin{figure*}
\centering
\includegraphics[width=0.9\textwidth]{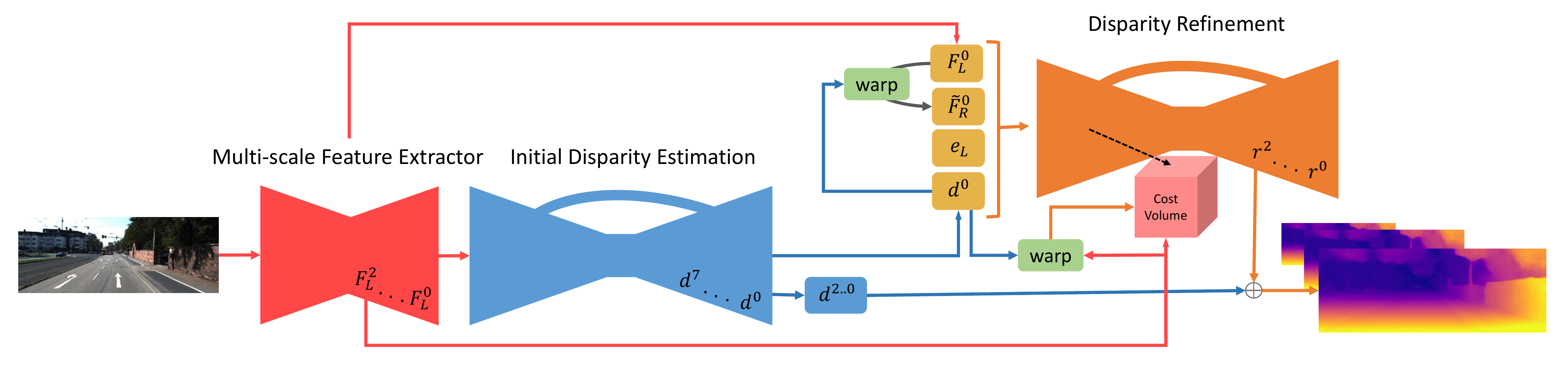}
\caption{Illustration of our \textit{monoResMatch} architecture. Given one input image, the multi-scale feature extractor (in red) generates high-level representations in the first stage. The initial disparity estimator (in blue) yields multi-scale disparity maps aligned with the left and right frames of a stereo pair. The disparity refinement module (in orange) is in charge of refining the initial left disparity relying on features computed in the first stage, disparities generated in the second stage, matching costs between high-dimensional features $F_L^0$ extracted from input and synthetic $\tilde{F}_R^0$ from a \emph{virtual} right viewpoint, together with absolute error $e_L$ between $F_L^0$ and back-warped $\tilde{F}_R^0$ (see Section \ref{sec:refinement}). }
\label{fig:monoResMatch}
\end{figure*}

%% file: figures/proxy_example.tex
\begin{figure*}
    \centering
    \renewcommand{\tabcolsep}{1pt}   
    \begin{tabular}{ccc}

        \includegraphics[width=0.3\textwidth]{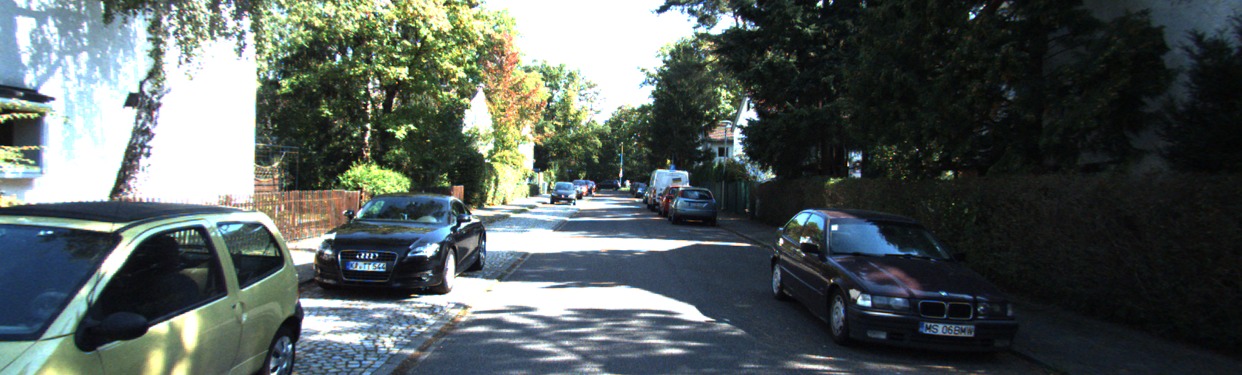}  &
        \includegraphics[width=0.3\textwidth]{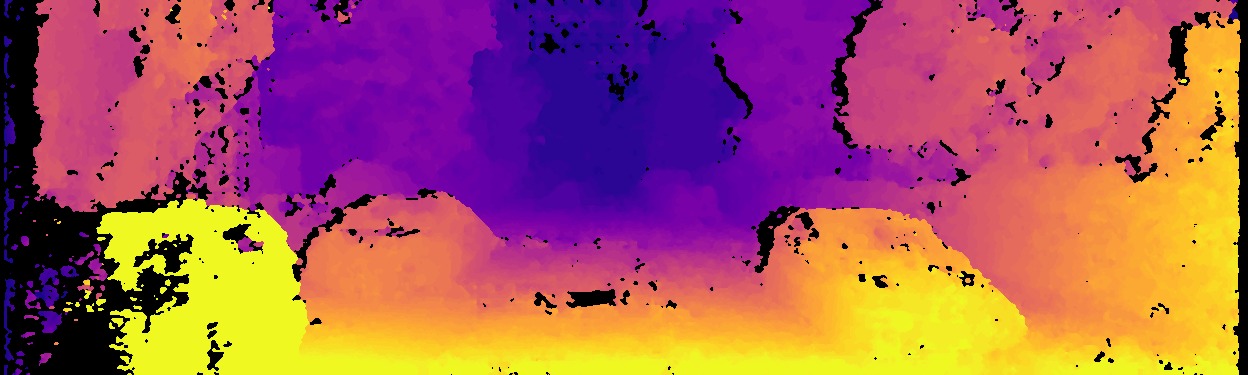}  &
        \includegraphics[width=0.3\textwidth]{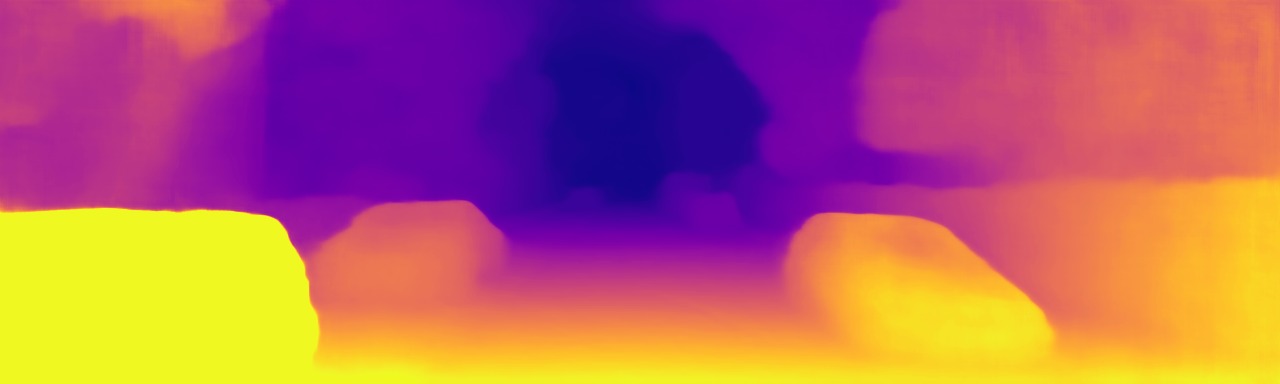} 
        \\
        \includegraphics[width=0.3\textwidth]{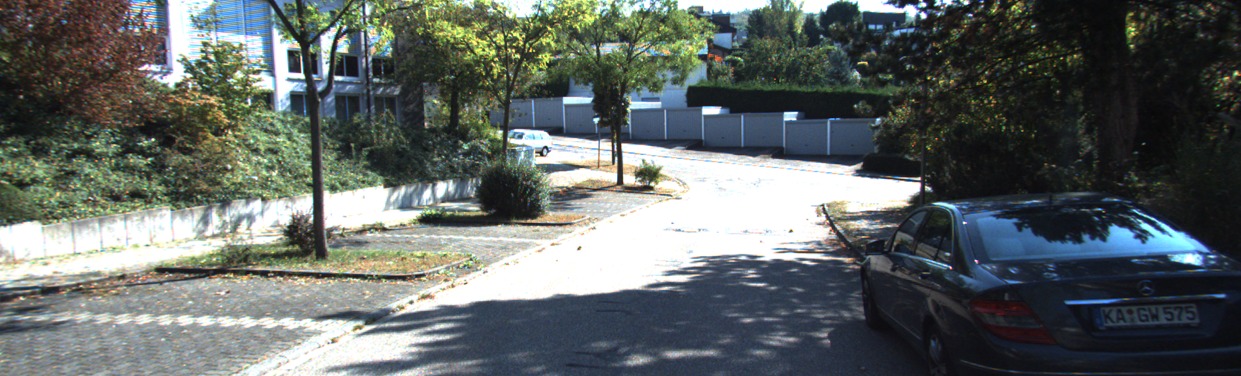}  &
        
        \includegraphics[width=0.3\textwidth]{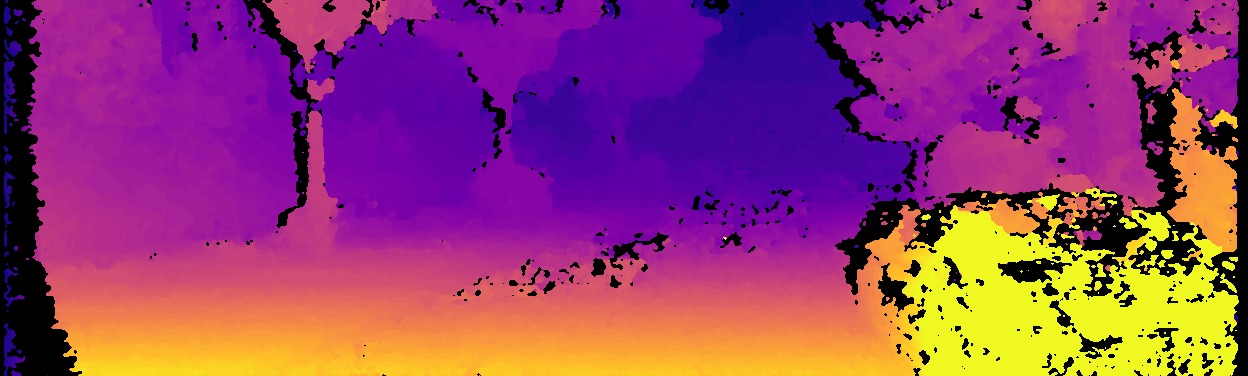} &
        \includegraphics[width=0.3\textwidth]{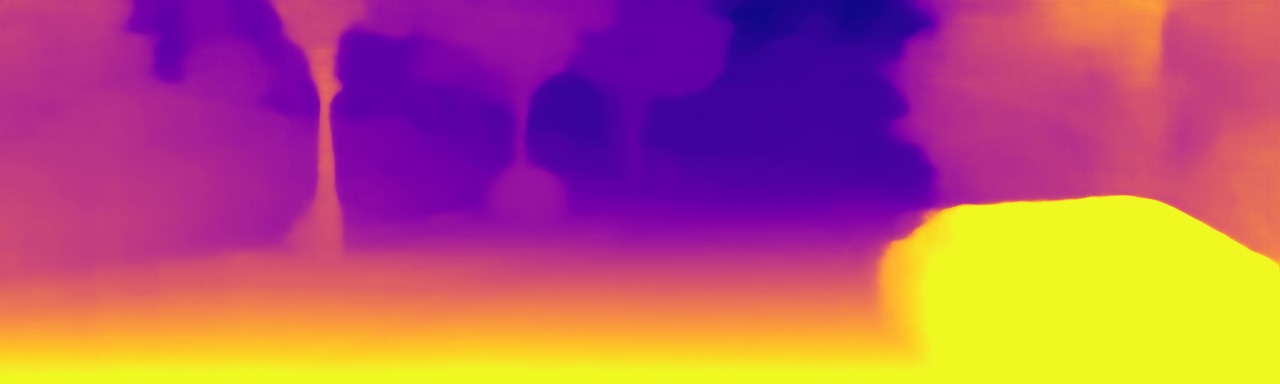}  
        \\
        (a) & (b) & (c)\\
    \end{tabular}
    \caption{Examples of proxy labels computed by SGM. Given the source image (a), the network exploits the SGM supervision filtered with left-right consistency check (b) in order to train monoResMatch to estimate the final disparity map (c). No post-processing from \cite{monodepth17} is performed on (c) in this example.}
    \label{fig:distillation}
\end{figure*}

%% file: equations/main_loss.tex
\begin{equation} \label{eq:initial_loss}
\begin{split}
	\mathcal{L}_{init} = & \alpha_{ap} (\mathcal{L}_{ap}^l + \mathcal{L}_{ap}^r) + \alpha_{ds} (\mathcal{L}_{ds}^l  + \mathcal{L}_{ds}^r)\\ 
	& + \alpha_{ps} (\mathcal{L}_{ps}^l + \mathcal{L}_{ps}^r)
\end{split}
\end{equation}

\begin{equation} \label{eq:initial_refined}
	\mathcal{L}_{ref} = \alpha_{ap} \mathcal{L}_{ap}^l + \alpha_{ds} \mathcal{L}_{ds}^l + \alpha_{ps} \mathcal{L}_{ps}^l
\end{equation}

%% file: equations/reconstruction_loss.tex
\begin{equation} \label{eq:image_rec}
	\mathcal{L}_{ap} = \frac{1}{N}\sum_{i,j}\alpha \frac{1-SSIM(I_{ij},\hat{I}_{ij})}{2} +  (1-\alpha) |I_{ij}-\hat{I}_{ij}|
\end{equation}

%% file: equations/smoothness_loss.tex
\begin{equation}\label{eq:smooth}
\mathcal{L}_{ds} = \frac{1}{N} \sum_{i,j}|\partial_{x} d_{ij}| e^{-|\partial_{x}I_{ij} |} + |\partial_{y} d_{ij}| e^{-|\partial_{y}I_{ij}|}
\end{equation}

%% file: equations/berhu_loss.tex
\begin{equation}\label{eq:semisupervised}
\mathcal{L}_{ps} = \frac{1}{N} \sum_{i,j} berHu(d_{ij},d_{ij}^{st}, c)
\end{equation}

\begin{equation}\label{eq:berhu}
    berHu(d_{ij},d_{ij}^{st},c)=
\begin{cases}
|d_{ij}-d_{ij}^{st}| & \text{if } |d_{ij}-d_{ij}^{st}|\leq c \\
\frac{|d_{ij}-d_{ij}^{st}|^2 - c^2}{2c} & \text{otherwise } \\
\end{cases}
\end{equation}

%% file: tables/godard_comparison.tex
\begin{table*}[!h] 
\centering
\scalebox{0.80}{
\begin{tabular}{lccc|cccc|ccc}
 \multicolumn{3}{c}{} & \multicolumn{3}{c}{} &\multicolumn{2}{c}{\cellcolor{blue!25} Lower is better}
 & \multicolumn{2}{c}{\cellcolor{LightCyan} Higher is better} \\
\hline
\toprule
Method & \multicolumn{2}{c}{Supervision} &
Train set & \cellcolor{blue!25} Abs Rel & \cellcolor{blue!25} Sq Rel & \cellcolor{blue!25} RMSE & \cellcolor{blue!25} RMSE log &  \cellcolor{LightCyan}$\delta<$1.25 &  \cellcolor{LightCyan}$\delta<1.25^2$ & \cellcolor{LightCyan}$\delta<1.25^3$  \\
\cmidrule(l){2-3} & Image & SGM \\
\midrule

Godard \etal \cite{monodepth17} ResNet50 & \checkmark &  & K & 0.128 & 1.038 & 5.355 & 0.223  & 0.833 & 0.939 & 0.972\\

Poggi \etal \cite{3net18} ResNet50 & \checkmark &  & K & 0.126 & 0.961 & 5.205 & 0.220  & 0.835 & 0.941 & 0.974\\



\textbf{monoResMatch} & \checkmark &  & K & 0.116 & 0.986 & 5.098 & 0.214  &  0.847 & 0.939 & 0.972 \\


\textbf{monoResMatch} & \checkmark & \checkmark & K & \textbf{0.111} & \textbf{0.867} & \textbf{4.714} &  \textbf{0.199} & \textbf{0.864} & \textbf{0.954} & \textbf{0.979} \\
\hline
Godard \etal \cite{monodepth17} ResNet50 & \checkmark & & CS,K & 0.114 & 0.898 & 4.935 & 0.206 & 0.861 & 0.949 & 0.976 \\
Poggi \etal \cite{3net18} ResNet50 & \checkmark & & CS,K & 0.111 & 0.849 & 4.822 & 0.202 & 0.865 & 0.952 & 0.978 \\
Godard \etal \cite{monodepth17} ResNet50 &  \checkmark & \checkmark & CS,K & 0.110 & 0.822 & 4.675 & 0.199 & 0.862 & 0.953 & 0.980 \\
\textbf{monoResMatch} (no-refinement) & \checkmark & \checkmark & CS,K & 0.107 & 0.781 & 4.588 & 0.195  & 0.869 & 0.957 & 0.980 \\
\textbf{monoResMatch} (no-corr) & \checkmark & \checkmark & CS,K & 0.104 & 0.766 & 4.553 & 0.192  & 0.875 & 0.958 & 0.980 \\
\textbf{monoResMatch} (no-pp) & \checkmark & \checkmark & CS,K & 0.098 & 0.711 & 4.433 & 0.189 & 0.888 & 0.960 & 0.980 \\
\textbf{monoResMatch}  & \checkmark & \checkmark & CS,K & \textbf{0.096} & \textbf{0.673} & \textbf{4.351} &  \textbf{0.184} & \textbf{0.890} & \textbf{0.961} & \textbf{0.981} \\

\hline

\end{tabular}
}
\smallskip
\caption{Ablation studies on the Eigen split \cite{eigen2014depth}, with maximum depth set to 80m. All networks run post-processing as in \cite{monodepth17} unless otherwise specified.}
\label{tab:comparison_monodepth}
\end{table*}

%% file: tables/overall_comparison.tex
\begin{table*}[!htbp]
\centering
\scalebox{0.80}{
\begin{tabular}{l|c|c|cccc|ccc}
\multicolumn{5}{c}{} &\multicolumn{2}{c}{\cellcolor{blue!25} Lower is better}
 & \multicolumn{2}{c}{\cellcolor{LightCyan} Higher is better} \\
\hline
Method & Supervision & Train set & \cellcolor{blue!25} Abs Rel & \cellcolor{blue!25} Sq Rel & \cellcolor{blue!25} RMSE & \cellcolor{blue!25} RMSE log &  \cellcolor{LightCyan}$\delta<$1.25 &  \cellcolor{LightCyan}$\delta<1.25^2$ & \cellcolor{LightCyan}$\delta<1.25^3$ \\
\hline

Zou \etal \cite{zou2018df} & Seq & CS,K & 0.146 &  1.182 &  5.215 &  0.213 & 0.818 & 0.943 & 0.978 \\
Mahjourian \etal \cite{mahjourian2018unsupervised} & Seq & CS,K & 0.159 & 1.231 & 5.912 & 0.243 & 0.784 & 0.923 &  0.970 \\
Yin \etal \cite{yin2018geonet} GeoNet ResNet50 & Seq & CS,K & 0.153 & 1.328 & 5.737 & 0.232 & 0.802 & 0.934 & 0.972\\
Wang \etal \cite{wang2018unsupervised} & Seq & CS,K & 0.148 & 1.187 & 5.496 & 0.226 & 0.812 & 0.938 & 0.975\\
Poggi \etal \cite{pydnet18} PyD-Net (200) & Stereo & CS,K & 0.146 & 1.291 & 5.907 & 0.245 & 0.801 & 0.926 & 0.967 \\
Godard \etal \cite{monodepth17} ResNet50 & Stereo & CS,K & 0.114 & 0.898 & 4.935 & 0.206 & 0.861 & 0.949 & 0.976 \\
Poggi \etal \cite{3net18} 3Net ResNet50  & Stereo & CS,K & 0.111 & 0.849 & 4.822 & 0.202 & 0.865 & 0.952 & 0.978 \\
Pilzer \etal \cite{Pilzer_2019_CVPR} (Teacher)  & Stereo & CS,K & 0.098 & 0.831 & 4.656 & 0.202 & 0.882 & 0.948 & 0.973 \\

Yang \etal \cite{yang2018deep} & Seq+Stereo & K$_o$, K$_{r}$, K$_{o}$ & 0.097 & 0.734 & 4.442 &  0.187 &  0.888 & 0.958 & 0.980 \\
\textbf{monoResMatch}  & Stereo & CS,K & \textbf{0.096} & \textbf{0.673} & \textbf{4.351} &  \textbf{0.184} & \textbf{0.890} & \textbf{0.961} & \textbf{0.981} \\

\hline
\end{tabular}
}
\smallskip
\caption{Quantitative evaluation on the test set of KITTI dataset \cite{KITTI_RAW} using the split of Eigen \etal \cite{eigen2014depth}, maximum depth: 80m. Last four entries include post-processing \cite{monodepth17}. K$_o$, K$_{r}$, K$_{o}$ are splits from K, defined in \cite{yang2018deep}. Best results are shown in bold. }
\label{table:eigen}
\end{table*}

%% file: tables/luo_comparison.tex
\begin{table*}[!ht]
\centering
\scalebox{0.80}{
\begin{tabular}{lccccc|cccc|ccc}
\multicolumn{8}{c}{} &\multicolumn{2}{c}{\cellcolor{blue!25} Lower is better}
 & \multicolumn{2}{c}{\cellcolor{LightCyan} Higher is better} \\
\hline
\toprule
Method & \multicolumn{5}{c}{Supervision} &
\cellcolor{blue!25}Abs Rel & \cellcolor{blue!25} Sq Rel & \cellcolor{blue!25} RMSE & \cellcolor{blue!25} RMSE log &  \cellcolor{LightCyan}$\delta<$1.25 &  \cellcolor{LightCyan}$\delta<1.25^2$ & \cellcolor{LightCyan}$\delta<1.25^3$  \\
\cmidrule(l){2-6} & 200-acrt & 100 &  200 & 500 & 700 \\
\midrule

Luo \etal \cite{luo2018single} & \checkmark  &   &  & &     & 0.101 & 0.673 & 4.425 & \textbf{0.176} & - & - & - \\
\textbf{monoResMatch} & \checkmark  &  &  & &  & \textbf{0.089} & \textbf{0.575} & \textbf{4.186} & 0.181  & 0.897 & 0.964 & 0.982 \\ 
\hline
Luo \etal \cite{luo2018single} &  \checkmark &  & \checkmark &   &   & 0.100 & 0.670 & 4.437 & 0.192 & 0.882 & 0.958 & 0.979 \\ 
\textbf{monoResMatch} &  \checkmark &  & \checkmark &   &   & \textbf{0.096} & \textbf{0.573} & \textbf{3.950} &  \textbf{0.168} &  \textbf{0.897} & \textbf{0.968} & \textcolor{red}{\textbf{0.987}} \\
\hline
Luo \etal \cite{luo2018single} & \checkmark &  &   & \checkmark  &   & 0.094 & 0.635 & 4.275 & 0.179 & 0.889 & 0.964 & 0.984 \\ 
\textbf{monoResMatch} & \checkmark &  &  & \checkmark  &   & \textcolor{red}{\textbf{0.093}} & \textcolor{red}{\textbf{0.567}} & \textcolor{red}{\textbf{3.914}} & \textcolor{red}{\textbf{0.165}} & \textcolor{red}{\textbf{0.901}} & \textcolor{red}{\textbf{0.969}} & \textcolor{red}{\textbf{0.987}} \\
\hline
Luo \etal \cite{luo2018single} & \checkmark &   &  & &  \checkmark   & \textbf{0.094} & 0.626 & 4.252 & 0.177 & 0.891 & 0.965 & 0.984 \\
\textbf{monoResMatch} &  \checkmark &  &  &   & \checkmark  & 0.095 & \textcolor{red}{\textbf{0.567}} & \textbf{3.942} & \textbf{0.166} & \textbf{0.899} & \textcolor{red}{\textbf{0.969}} & \textcolor{red}{\textbf{0.987}} \\
\hline
Guo \etal \cite{guo2018learning} &  & \checkmark  &  &  &   & \textbf{0.096} & 0.641 & 4.095 & \textbf{0.168} & 0.892 & 0.967 & 0.986 \\

\textbf{monoResMatch} &  & \checkmark &  & &  & 0.098 & \textbf{0.597} & \textbf{3.973} & 0.169 & \textbf{0.895} & \textbf{0.968} & \textcolor{red}{\textbf{0.987}} \\ 

\hline
\end{tabular}
}
\smallskip
\caption{Experimental results on the Eigen split \cite{eigen2014depth}, maximum depth: 80m. Comparison between methods supervised by few annotated samples. Best results in direct comparisons are shown in bold, best overall scores are in red, consistently attained by monoResMatch.}
\label{table:eigen_200_500_700}
\end{table*}

%% file: figures/kitti_benchmark.tex
\begin{figure*}[t!]
    \centering
        \includegraphics[width=2.22in]{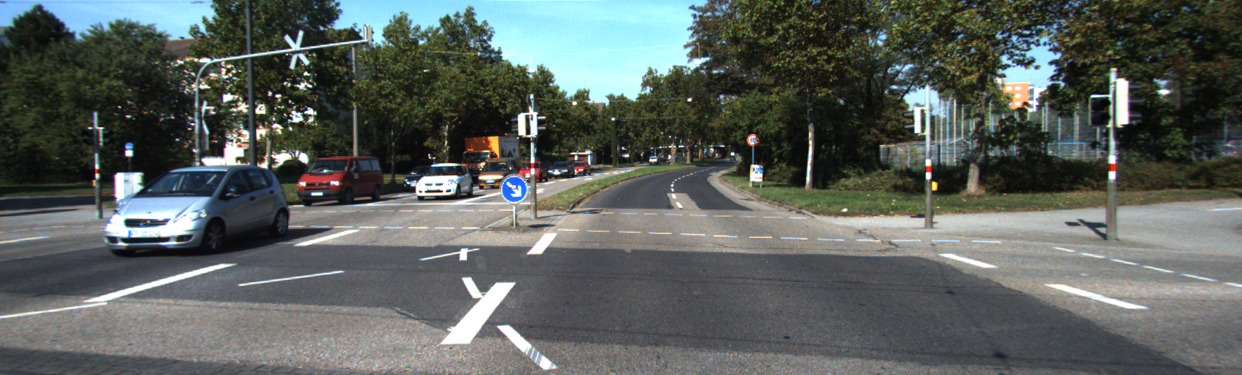}
        \includegraphics[width=2.22in]{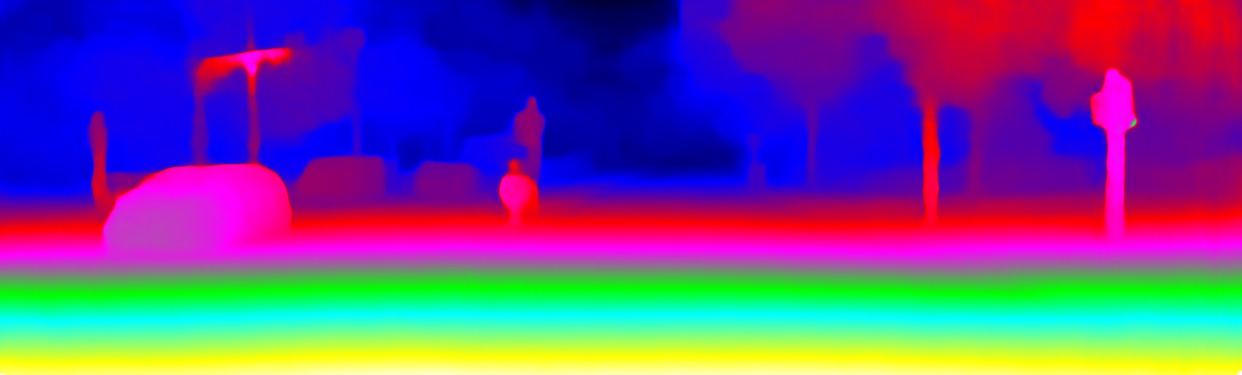}
        \includegraphics[width=2.22in]{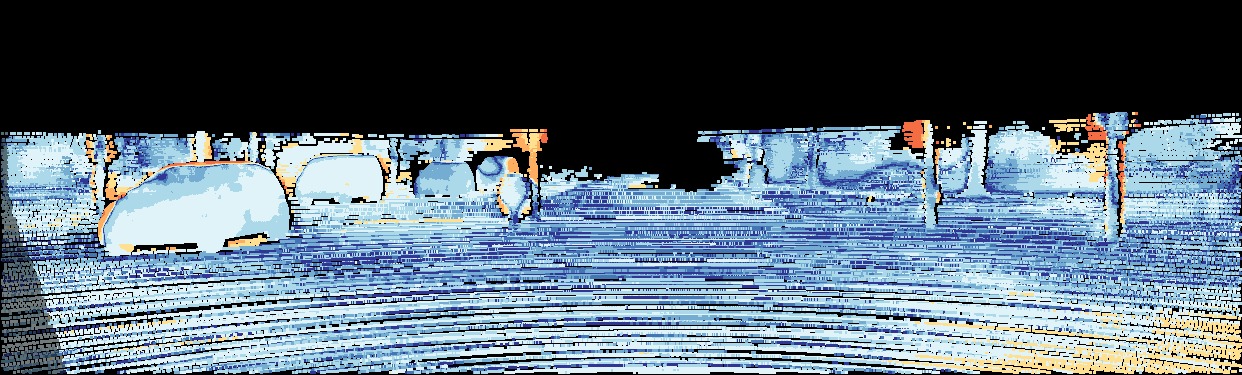}
        \includegraphics[width=2.22in]{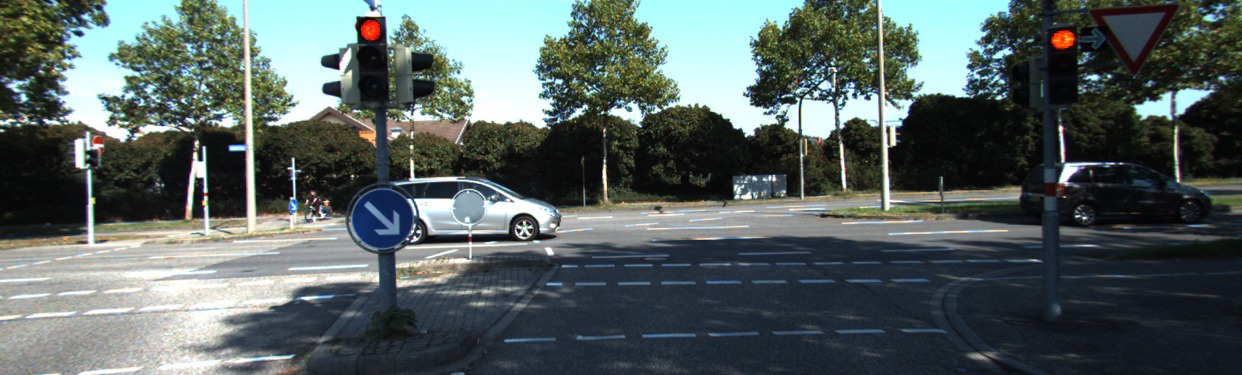}
        \includegraphics[width=2.22in]{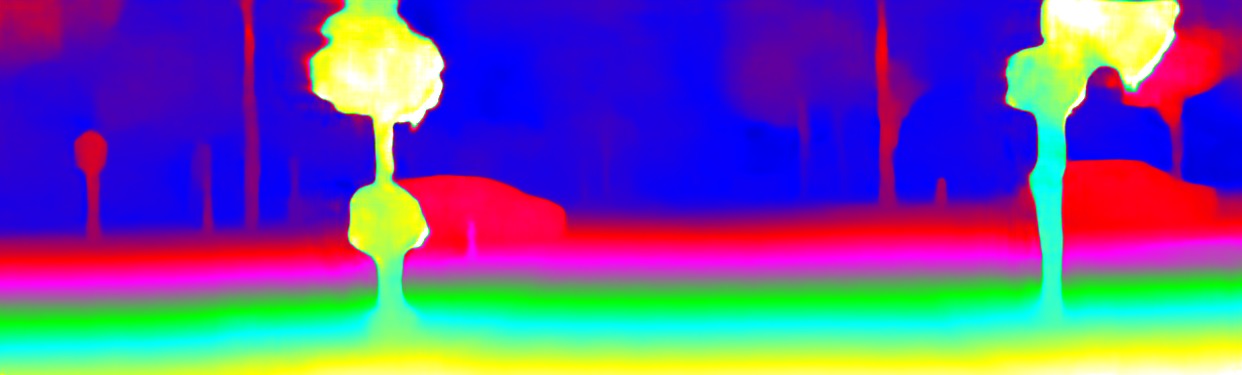}
        \includegraphics[width=2.22in]{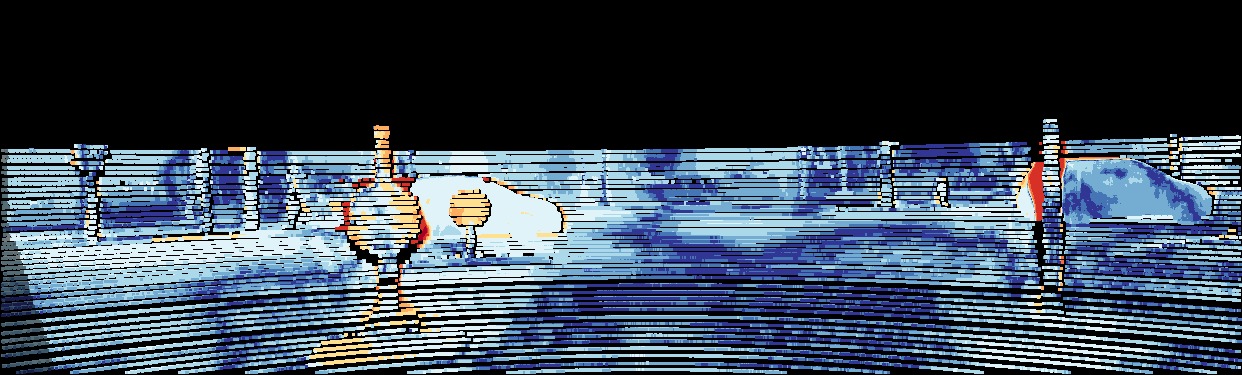}
        \includegraphics[width=2.22in]{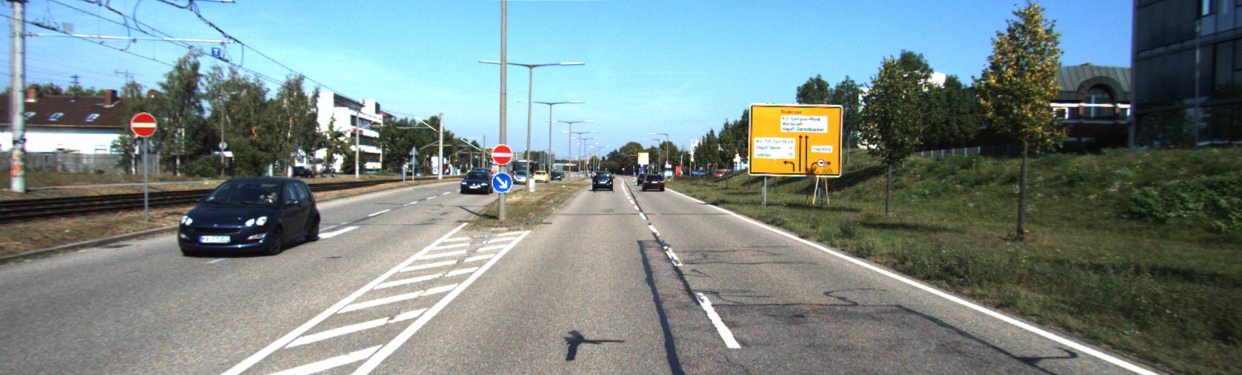}
        \includegraphics[width=2.22in]{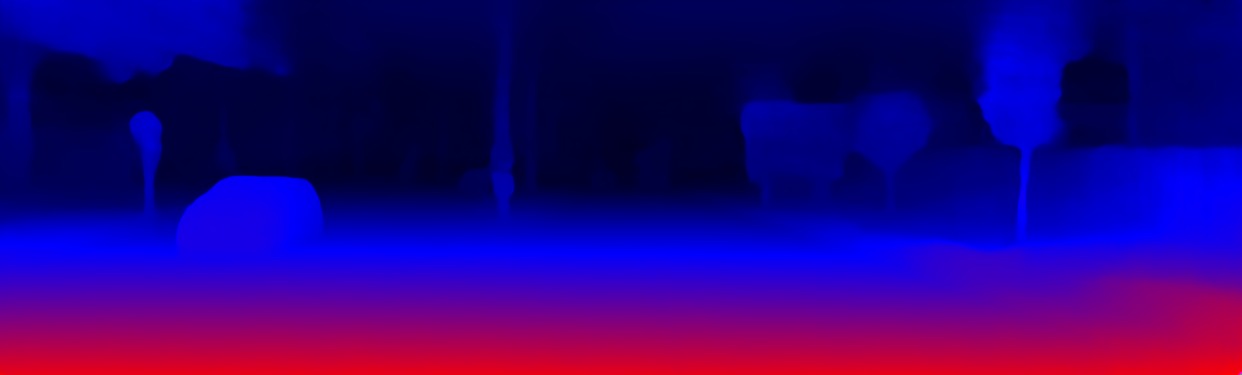}
        \includegraphics[width=2.22in]{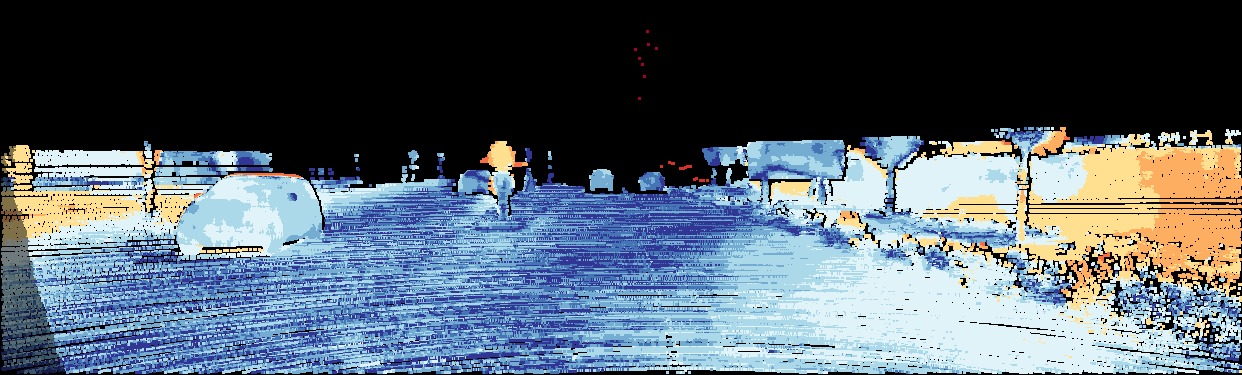}
         \includegraphics[width=0.98\textwidth]{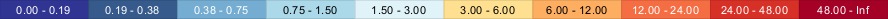}
    \caption{Stereo evaluation of our depth-from-mono framework. From left to right the input image, the predicted depth and the errors with respect to ground truth. The last line reports the color code used to display the seriousness of the shortcomings (same of \cite{KITTI_2015})}
    \label{fig:kitti_test}
\end{figure*} 

%% file: tables/kitti_benchmark.tex
\begin{table}
\centering
\begin{tabular}{|l|c|c|c|}
\hline
Method       & D1-bg & D1-fg & D1-all \\ \hline
monodepth \cite{monodepth17} & 27.00  & 28.24 & 27.21  \\ \hline
OCV-BM       & 24.29 & 30.13 & 25.27  \\ \hline
SVS \cite{luo2018single} & 25.18 & 20.77 & 24.44 \\ \hline
\textbf{monoResMatch}  & \textbf{22.10} & \textbf{19.81} & \textbf{21.72} \\ \hline
\end{tabular}
\smallskip
\caption{Quantitative results on the test set of the KITTI 2015 Stereo Benchmark \cite{KITTI_2015}. Percentage of pixels having error larger than 3 or 5\% of the ground truth. Best results are shown in bold. }
\label{table:kitti-stereo}
\vspace{-10pt}
\end{table}